\documentclass[10pt,twocolumn,letterpaper]{article}

\usepackage{3dv}
\usepackage{times}
\usepackage{epsfig}
\usepackage{graphicx}
\usepackage{amsmath}
\usepackage{amssymb}
\usepackage{algpseudocode}
\usepackage{algorithm}
\usepackage{balance}

\usepackage{setspace}

\usepackage[pagebackref=true,breaklinks=true,colorlinks,bookmarks=false]{hyperref}

\threedvfinalcopy 

\def\threedvPaperID{278} 

\ifthreedvfinal\fi
\begin{document}

\title{3D Lip Event Detection via Interframe Motion Divergence at Multiple Temporal Resolutions}

\author{Jie Zhang\\
School of Artificial Intelligence, Beijing Technology and Business University, China\\
{\tt\small zhangjie09@buaa.edu.cn}
\and
Robert B. Fisher\\
School of Informatics, The University of Edinburgh,UK\\
{\tt\small rbf@inf.ed.ac.uk}
}

\maketitle
\thispagestyle{empty}

\begin{abstract}
The lip is a dominant dynamic facial unit when a person is speaking.
Detecting lip events is beneficial to speech analysis and support
	for the hearing impaired. This paper proposes a 3D lip event detection
	pipeline that automatically determines the lip events from a 3D
	speaking lip sequence. We define a motion divergence measure
	using 3D lip landmarks to quantify the interframe dynamics
	of a 3D speaking lip. Then, we cast the interframe motion
	detection in a multi-temporal-resolution framework that allows
	the detection to be applicable to different speaking speeds.
	The experiments on the S3DFM Dataset investigate the overall
	3D lip dynamics based on the proposed motion divergence. The
	proposed 3D pipeline is able to detect opening and closing
	lip events across 100 sequences, achieving a state-of-the-art performance.

\end{abstract}

\section{Introduction}
Speaking is a spontaneous behavior involving multiple biological modalities, including voice, visual speech \cite{FengVisual18}, dynamic face motions \cite{ZhangF19}, etc. There are many applications related with speaking, e.g. speech recognition, lip-reading \cite{Chung2017Lip}, or identity recognition \cite{FengVisual18, ZhangF19}. The lip is a dominant dynamic facial unit when speaking, and the visual lip acts as an important counterpart to the audio information. For visual lip related applications, detecting the lip events of opening and closing is a significant prerequisite for analyzing lip behavior.

Lip event detection aims at localizing and tracking the lip region across a video sequence and then determining the starting and ending times of the speaking behavior along the temporal domain. The core of lip event analysis focuses on the spatio-temporal representation of interframe motion, instead of the whole lip dynamics. The challenge always lies at fine-grained temporal detection (i.e. frame-level motion decision).

Existing lip event analysis algorithms are mainly based on 2D image sequences. A group of lip event detection methods are based on dense motion field analysis. For example, Karlsson and Bigun \cite{lipevent12} improved 2D optical flow estimation and constructed a low-level lip dynamics feature for lip event detection. Liu et al. \cite{LiuCT16} constructed oriented histograms of regional optical flow (OH-ROF) over 2D lip sequences to represent frame-level lip motion, and then proposed a low-rank affinity pursuit approach to determine the starting and ending of a lip event. This method is efficient as it is free of prior learning.

There are two drawbacks to using 2D motion fields: 1) the inner mouth is cluttered and its motion field affects the lip event detection; 2) the motion field of sequential 2D intensity images is sensitive to facial pose variations. Another family of methods focuses on lip shape deformation \cite {1660092,7099240} and motion features \cite {song2014visual,7420734}. Taeyup et al. \cite{song2014visual} proposed a chaos-like lip motion measure - fractal trajectories observed in phase space, which is especially robust against illumination changes. Patrona et al. \cite{7420734} utilized multiple intensity image features and dense trajectories of keypoints to represent both the local shape and motion of a speaking face. The representations are then integrated into a bag of words model for later classification. Recently, an end-to-end network - HiCA \cite{8803248} was designed to extract the local and global temporal features and achieves visual voice activity detection. There are also some dual-modality approaches \cite{dov2015audio,ariav2018deep,8649655} that combine audio and intensity video information for speaking event detection. The two modalities complement each other. Overall, lip event detection via 2D sequences always suffers from the common drawbacks of intensity images. It is sensitive to facial pose variations, nonuniform illumination, scale changes, etc.

Lip event detection via 3D sequence data is a promising alternative approach for the task. To the best of our knowledge, 3D lip event analysis is a less investigated approach. 3D lip event detection is challenging when dealing with 3D noisy data, cluttered backgrounds, and frame-level non-rigid deformation. Since the 3D mouth cavity is usually reconstructed with lower quality due to the darkness and occlusion of oral components, the 3D motion field based approach is not a preferred solution. Besides, speakers with low speaking speeds generate weak deformations between consecutive frames, which are tough to be detected and compared.

To tackle the above problems, we focus on 3D lip event detection based on the interframe motion of 3D lip landmarks. The landmark-based motion representation is more immune to the cluttered and dark background of the mouth cavity. The main contributions of this paper are:
\begin{itemize}
  \item We define a new interframe motion representation for the 3D speaking lip - 3D motion divergence. The motion signature quantifies the overall deformation of 3D lip landmarks via a reference sphere over the 3D lip. (Sec. \ref{subsecmotion})
  \item We propose a new 3D lip event detection pipeline that determines the lip opening and closing frames from the 3D video in multiple temporal resolutions. The coarse-to-fine temporal strategy is beneficial when dealing with various motion speeds. (Sec. \ref{subsecdetection})
\end{itemize}
The proposed algorithms were verified on a 3D speaking face dataset (S3DFM \cite{ZhangF19,0051RF18})\footnote{http://groups.inf.ed.ac.uk/trimbot2020/DYNAMICFACES/} and has good detection performance over 100 sequences with 200 events.

\section{Overview of the proposed pipeline}
\label{sec_framework}
We define two lip events as 1) the moment lips first start to open and 2) the moment lips have finished closing.

\begin{figure*}
  \centering
  \includegraphics[width=15cm]{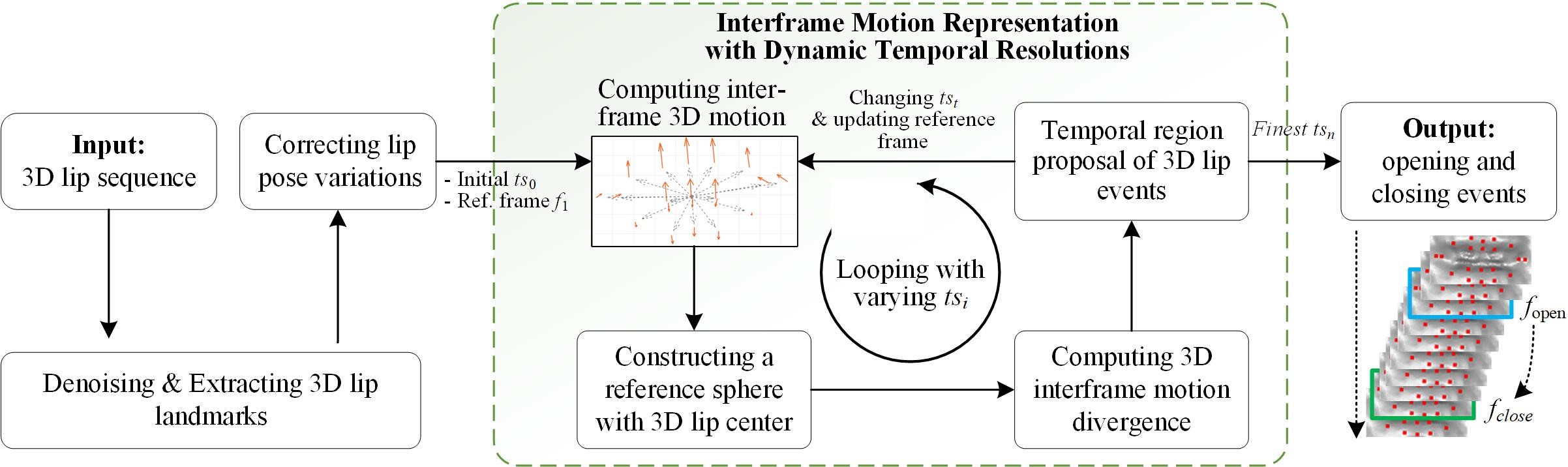}
  \caption{The proposed pipeline of 3D lip event detection. Given a 3D lip sequence, the lip event detection begins with a coarse temporal resolution $ts_0$ and an event reference frame $f_1$, and it gradually refines the temporal region proposal by increasing temporal resolution $ts_i$ as well as updating the reference frame. At each temporal resolution, we sequentially compute the interframe divergence and classify the frames.}\label{figframework}
\end{figure*}

The proposed pipeline for 3D lip event detection is shown in Fig.\ref{figframework}. Given a 3D lip video sequence, we first perform noise reduction to improve the overall quality of the 3D data. Then, the 3D landmarks of the dynamic lips are extracted using a non-rigid 3D registration algorithm with a 3D deformable model \cite{AmbergRV07,GerigMBELSV18}. The registration also facilitates the rigid pose correction of the lip. The motions of lip landmarks are fed into the proposed interframe motion representation, following a framework of dynamic temporal resolutions. The event region proposal is gradually refined to an event frame. More details are given in the next section.

\section{Interframe 3D lip motion representation}
\label{secrepre}
\subsection{Preprocessing 3D lip sequence}
\label{subsecpreprocessing}
The preprocessing incorporates two main steps: denoising and rigid pose correction. The raw 3D point cloud sequences usually suffer from some spatial noise and temporal fluctuations, due to the sensor technology and data capture procedure. To improve the overall quality of the 3D data, we firstly denoise the 3D point cloud sequence using a multi-frame fusion algorithm \cite{noiseJVCI18}, but do not reduce the frame rate. On the other hand, facial pose is likely to slightly change while a person is speaking. The rigid lip pose variation mixed with non-rigid lip deformation will affect the interframe lip motion analysis. The lip landmarks extraction involves rigid registration that is used for correcting the lip pose.


\subsection{Interframe 3D lip motion signature}
\label{subsecmotion}
It is interesting to note that the lip events of opening and closing are not the same for every person and every phrase that the speaker is about to pronounce. According to the research on lip motion-based behaviometrics \cite{BenediktCRM10,Hmm14,ZhangF19}, the lip event is a person-specific trait. Besides, the motion vector of an individual lip landmark cannot represent the opening or closing state of a lip. E.g. an opening lip can contain both diverging and converging lip landmarks. Based on the above observations, the interframe motion signature of a 3D speaking lip should capture common and global properties of the frame-level 3D lip dynamics, and cope with both person-specific and syllable-specific differences. Another common property of the 3D lip dynamics while speaking is the symmetry, which means that speaking is a regular and text-constrained motion.

For raw 3D lip motion data, we only focus on the landmarks of the 3D lip, instead of a dense point-wise motion field. This is to avoid the impact of unstable motion from the inner mouth. Thus, we represent a 3D lip as a set of 3D landmarks $\{{\bf{L}}^t_i \in R^3\}(i = 1...n)$ at time $t$. Each lip landmark generates a motion vector ${\bf{m}}^t_i \in R^3$ during speaking events. Motivated by divergence in flow analysis, we define an interframe motion signature over all the 3D lip landmarks. The signature measures the energy and overall type of a lip motion between two frames. It is a global metric summarizing the interframe lip dynamics.

\begin{figure}
  \centering
  \includegraphics[width=7cm]{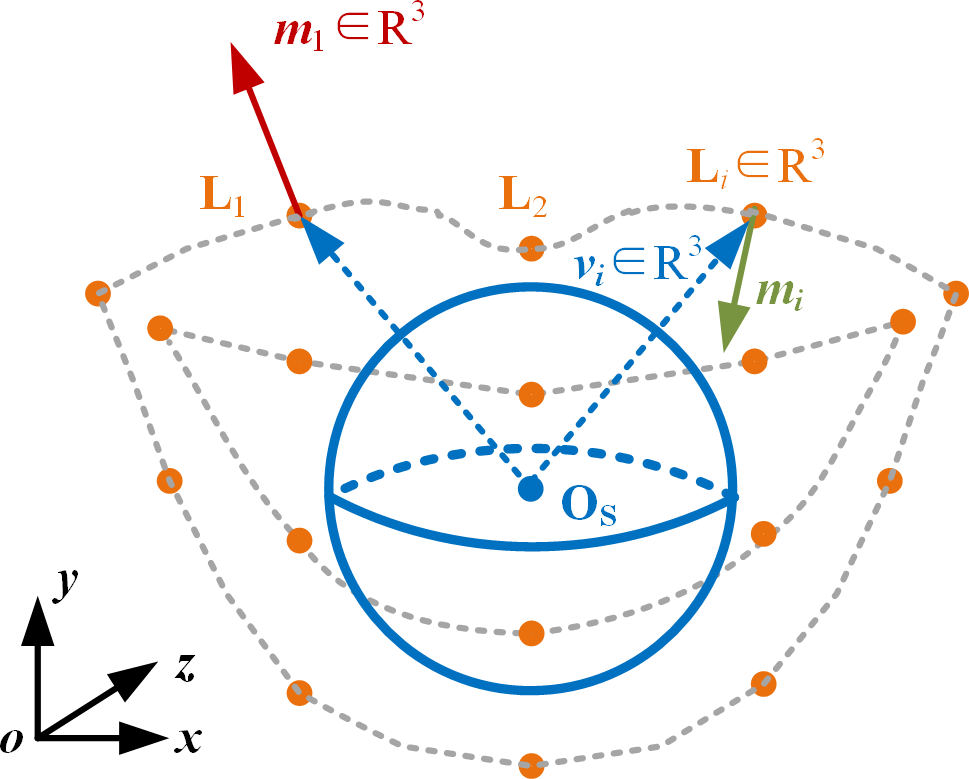}
  \caption{Reference sphere (blue ball) for calculating the interframe motion divergence of a 3D lip at one time. ${\bf{L}}_i$ is a 3D lip landmark. ${\bf{m}}_i$ and ${\bf{v}}_i$ is the 3D motion vector and reference vector of the $i_{th}$ lip landmark, respectively. The motion divergence can be calculated according to Eqn.2.}\label{figreference}
\end{figure}

\textbf{Lip motion signature energy}. For a lip event, we assume that the motion of every lip landmark brings a unit of ``energy'' into or out of the mouth region. To measure the motion energy of all the lip landmarks, we establish a reference sphere $S$ within the lip region in a 3D coordinate system $o-xyz$, as shown in Fig.\ref{figreference}. The center $O_S$ of the sphere $S$ is the 3D center of mass of all the lip landmarks at the reference frame. Since every frame has been registered with the reference frame before interframe motion analysis, the center is static across the sequence. The surface area of the sphere is a constant $V_S$. We focus on the ``energy'' change at the center $O_S$, which is defined precisely below.

For each point on the surface of the sphere, we define the radial direction as a unit normal motion direction ${\bf{v}}_i$. The motion vector at time $t$ and lip landmark $i$ is ${\bf{m}}^t_i={\bf{L}}^t_i-{\bf{L}}^{[t-1]}_i$. A set of motion vectors ${\bf{M}}^t\{{\bf{m}}^t_i \in R^3\}$ (as defined by the motion of lip landmark ${\bf{L}}^t_i$) can only generate the motion energy in the normal motion direction. Thus, the overall lip motion energy is defined as Eqn.\ref{eqndiv}.
\begin{equation}\label{eqndiv}
 Div_{lip}^{t} = \mathop {\lim }\limits_{S \to {O_S}} \frac{1}{{{V_S}}}\mathop{{\int\!\!\!\!\!\int}\mkern-21mu \bigcirc}\limits_\Omega
 {{{\bf{m}}^t_i}\cdot{{\bf{v}}_i}dS}
\end{equation}
where $\Omega$ is the closed surface of the sphere, with $dS$ is a unit of area. $V_S$ is the sphere's surface area. When all the energy of the motion converges at the center, i.e. $S \to {O_S}$, the divergence at the center measures the energy and type of the motion.

As the number $n$ of the lip landmarks is finite, we discretize the divergence of interframe lip motion at time $t$ as

\begin{equation}\label{eqndiscrete}
Div_{lip}^t = \frac{1}{{{V_S}}}\sum\limits_{i = 1}^n {{{\bf{m}}^t_i}\cdot \amalg({{\bf{L}}^t_i} - {{\bf{O}}^t_S})\Delta S}
\end{equation}
with
\begin{equation}\label{eqnarea}
V_S = Area(\Omega)  = \sum\limits_{i = 1}^n {\Delta S}
\end{equation}
where $\Delta S$ is a discrete fraction of the sphere's surface area, which is a constant. $\amalg(\cdot)$ is the vector normalization.

\textbf{Lip motion signature category}. Every interframe motion can be represented by the signature $Div_{lip}^t$. The event state of one lip landmark is defined by the angle between the 3D landmark motion vector ${\bf{m}}^t_i$ and its reference vector ${\bf{v}}_i=\amalg({{\bf{L}}^t_i} - {{\bf{O}}_S})$. When the angle is over 90 degrees, the landmark is closing, and vice versa. That is,
\begin{equation}
\label{eqnangle}
{\mathop{\rm sgn}} (Div_{lm}^t) = {\mathop{\rm sgn}} \langle {{\bf{m}}^t_i},{({\bf{L}}^t_i}-{\bf{O}}_S) \rangle
\end{equation}
where $sgn(\cdot)$ is a signum. $Div_{lm}^t$ is the motion divergence of a lip landmark. When the mouth is opening, the overall interframe motion divergence of all lip landmarks is a positive value and vice versa. We define the lip as static if the motion signature satisfies a threshold ${\varepsilon _{silence}}$, as Eqn.\ref{eqnsilence}. When the units of 3D lip points are in millimeters, we set ${\varepsilon _{silence}} =1$ in practice.
\begin{equation}\label{eqnsilence}
  \left| {Div^t_{lip}} \right| < {\varepsilon _{silence}}
\end{equation}
Otherwise
\begin{equation}\label{eqnstate}
\begin{array}{l}
LipState_t = \left\{ {\begin{array}{*{20}{c}}
{opening}&{{\mathop{\rm sgn}} (Di{v_{lip}^t}) =  1}\\
{closing}&{{\mathop{\rm sgn}} (Di{v_{lip}^t}) = -1}
\end{array}} \right.\\
\begin{array}{*{20}{c}}
{s.t.}
\end{array}\begin{array}{*{20}{c}}
{\begin{array}{*{20}{c}}
{\left| {Div_{left}^t - Div_{right}^t} \right| < \varepsilon _{symmetry}}
\end{array}}
\end{array}
\end{array}
\end{equation}
where $sgn(\cdot)$ is a signum. $Div_{left}^t$ and $Div_{right}^t$ are the divergences of the left side and right side of a lip at time $t$, respectively, and both satisfy Eqn.2. We incorporate the symmetry as another constraint in the lip event detection. That is, if the motion divergences of left lip $Div_{left}^t$ and right lip $Div_{right}^t$ are almost the same within a tolerance $\varepsilon _{symmetry}$, the interframe motion satisfies symmetry. Otherwise, the interframe motion will be rejected as a state of interest. In our implementation, the tolerance $\varepsilon _{symmetry}$ is set as 0.4 empirically.

The lip event is sequential and is defined as the first frame whose state is opening or the last frame whose state is closing.

\subsection{Event detection at multi-temporal-resolutions}
\label{subsecdetection}
\begin{figure}[H]
  \centering
  \includegraphics[width=8cm]{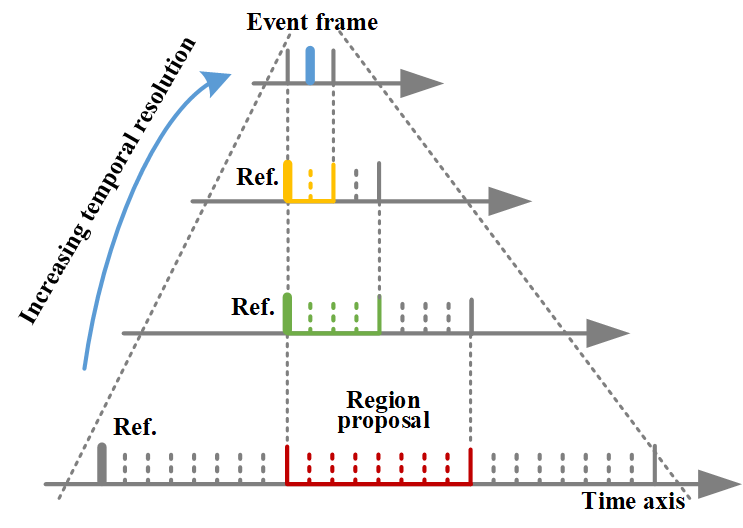}\\
  \caption{Lip event detection at multiple temporal resolutions: the detection begins at a coarse temporal resolution (averagely downsampling the original frames in the time domain), determine an interframe interval as a region proposal, increase the temporal resolution of the region by downsampling the original frames with smaller sampling interval, repeat the detection at each level of temporal resolution.}\label{fig_reso}
\end{figure}

We cast the interframe motion detection into a framework with multiple temporal resolutions, as shown in Fig. \ref{fig_reso}. As a temporal region with a coarse temporal resolution covers more motion energy, the pipeline begins with the detection at a coarse time scale (bottom axis in Fig. \ref{fig_reso}) and gradually increases the temporal detection resolution for fine detection (top axis in Fig. \ref{fig_reso}).

Given a 3D point cloud sequence, we define a coarse temporal resolution by averagely downsampling the original frames in the time domain. Assume that the initial temporal detection resolution is $ts_0\times$ frames. That is, the initial frame rate is $1/ts_0$ of the original frame rate. We set the first frame of the sequence as a reference frame. The interframe motion detection is sequentially performed along the time axis. The detection algorithm generates a temporal region proposal which includes the potential opening frame (or closing frame) of a lip event. Then, we update the reference frame as the first frame of the region and increase the temporal resolution by downsampling the original frames in the region proposal again (but with the downsampling interval smaller than that of the last round). The detection is conducted in each round. The region proposal is finally refined to one frame as the temporal detection resolution increases.

The merits of this hierarchical strategy are two-fold: 1) the coarse-to-fine detection hierarchically reduces the searching space for the target frame, which is beneficial to reduce the false detection rate; 2) coarse temporal resolution allows motion event detection for people with lower speaking speeds. When the speaking speed is slow, the interframe with the finest temporal resolution may not generate enough motion divergence for detection. The event response can be given at a coarser temporal resolution.

Finally, the proposed pipeline outputs the opening and closing frames of a 3D lip event. The overall algorithm is shown in Table \ref{alg:Framwork}.

 \begin{algorithm}[htb]
  \caption{Proposed pipeline of 3D lip event detection}
  \label{alg:Framwork}
  \begin{algorithmic}[1]
    \Require
      3D speaking face sequence $\{f_t\}(t=1...m)$;
    \Ensure
     Lip event frames $\{f_{open}, f_{close}\}$ (Note: we only present the procedure of lip opening detection as an example below, as the closing event detection is similar);
\State Initialize reference frame as $f_1$;
\State Construct a reference sphere at the 3D lip center;
\Procedure{Interframe Motion Representation}{}
\For{each temporal resolution $ts=\{30 \times, 15 \times, 7\times, 3\times, 1\times\}$ original interframe interval}
\For{each 3D frame $f_{ts\times j}(j=1 \cdots m/ts)$}
\State Denoise current 3D frame;
\State Register $f_{ts\times j}$ with reference frame;
\State Extract 3D lip landmarks $\{{\bf{L}}^{ts\times j}_{n}\}$;
\State Compute motion vectors of lip landmarks;
\State Compute 3D interframe motion divergence;
\State Determine event state of current lip;
\If{lip opening event happens}
\State Update reference frame with $f_{ts\times (j-1)}$;
\State Update event frame with $f_{open}=f_{ts\times j}$;
\State Break;
\EndIf
\EndFor
\EndFor
\EndProcedure\\
\Return $\{f_{open}, f_{close}\}$;
  \end{algorithmic}
\end{algorithm}

\section{Experiments and discussion}
\begin{figure}
  \centering
  \includegraphics[width=8.2cm]{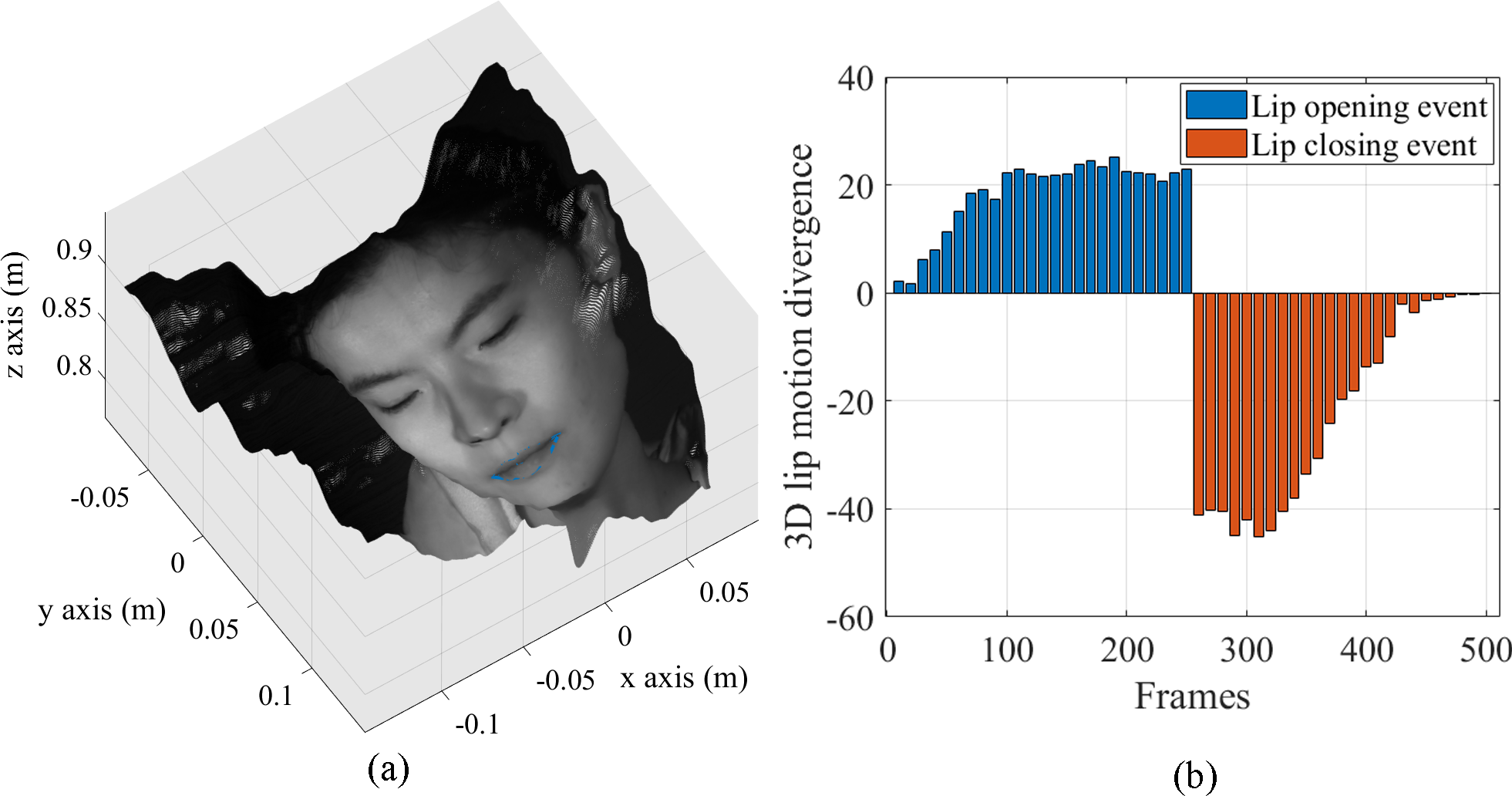}\\
  \caption{3D lip dynamics analysis: (a) an example 3D point cloud frame; (b) interframe motion divergence across a 3D speaking face sequence. The first 250 interframe motions are measured for opening detection, with the first frame as the reference frame. The last 250 interframe motions are used for closing detection, with the last frame as the reference frame.}\label{figdata}
\end{figure}
This section reports the experiments on a 3D speaking face dataset. We analyze the full lip dynamics of speaking and investigate the comparative performance of the proposed pipeline and some representative counterparts.

\subsection{Dataset}
The proposed pipeline was verified on a publicly available dynamic face dataset - Speech-driven 3D Facial Motion Dataset (S3DFM) \cite{S3DFM,0051RF18}. The dataset has multi-modality data from 77 subjects covering more than 20 nationalities. The facial dynamics is generated from the subject speaking a one-second short phrase ``ni'hao''. Each sample set contains a 2D intensity sequence $+$ a 3D point cloud sequence $+$ a synchronized audio sequence. The audio and video modalities of a lip event were collected with a light flash as a synchronization trigger. We only use the 3D speaking face modality in the algorithm presented here, while the audio and intensity modalities help determine the ground truth labeling. We set the frame rate of each sequence as 250 frames per second (fps) for better labeling. Each frame is a 3D point cloud with the resolution of 600 points $\times$ 600 points. An example 3D point cloud frame is shown in Fig.\ref{figdata}a.

We manually labelled the lip opening and closing frames using both the pixel-wise registered intensity sequences and the synchronized audio sequences. The audio clip firstly gave a coarse time localization and then we compared the consecutive frames around the coarse time to finally decide a fine ground-truth event frame.

\subsection{Qualitative analysis}
\label{sub_qualita}
\begin{figure*}[h]
  \centering
  \includegraphics[width=17cm]{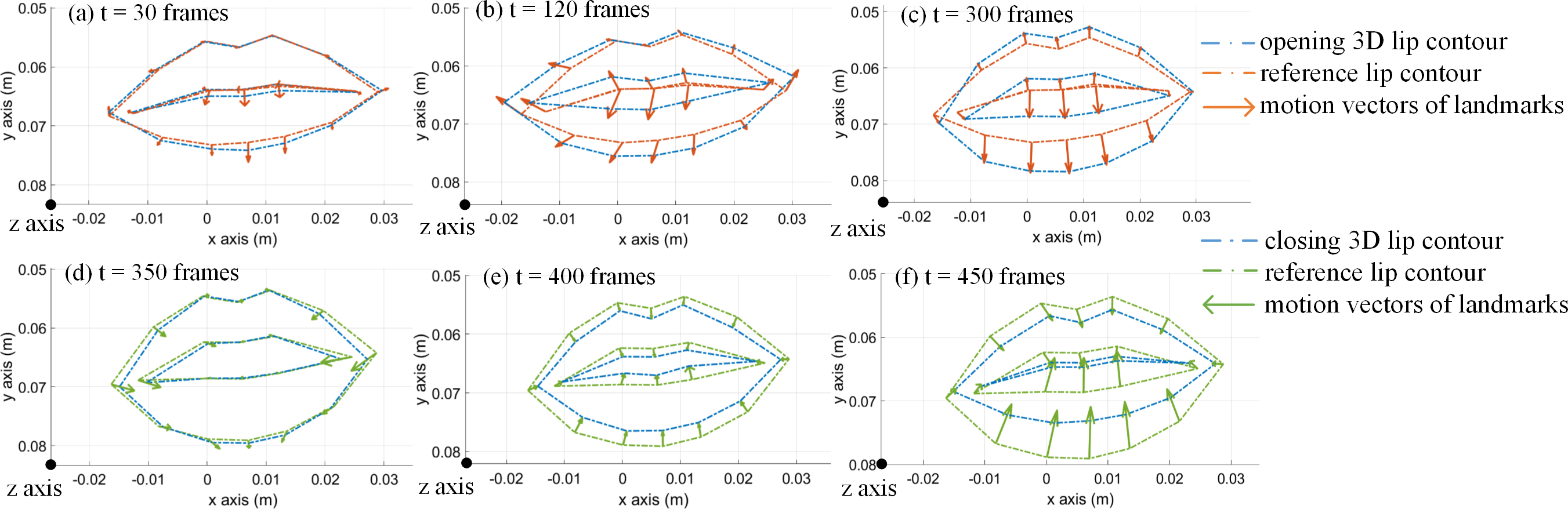}\\
  \caption{3D speaking lip motion in a 3D sequence of 500 frames: (a)-(c) opening frames at 3 moments; (d)-(f) closing frames at 3 moments.}
  \label{fig_motion}
\end{figure*}

\textbf{3D lip landmarks dynamics}. The 3D motion divergence of a whole sequence is shown in Fig.\ref{figdata}b, where we separate a whole sequence with 250 frames into the opening event region proposal of the first half frames and the closing event region proposal of the last half frames. We can see that the 3D motion divergence is increasing during the start of speaking (blue bars), and vice versa (red bars). Ax example of the dynamics of 3D lip landmarks across the whole sequence is shown in Fig.\ref{fig_motion}. The 3D motion vectors are diverging when the opening event happens, and they are converging in the closing event.

\begin{figure}
  \centering
  \includegraphics[width=8.5cm]{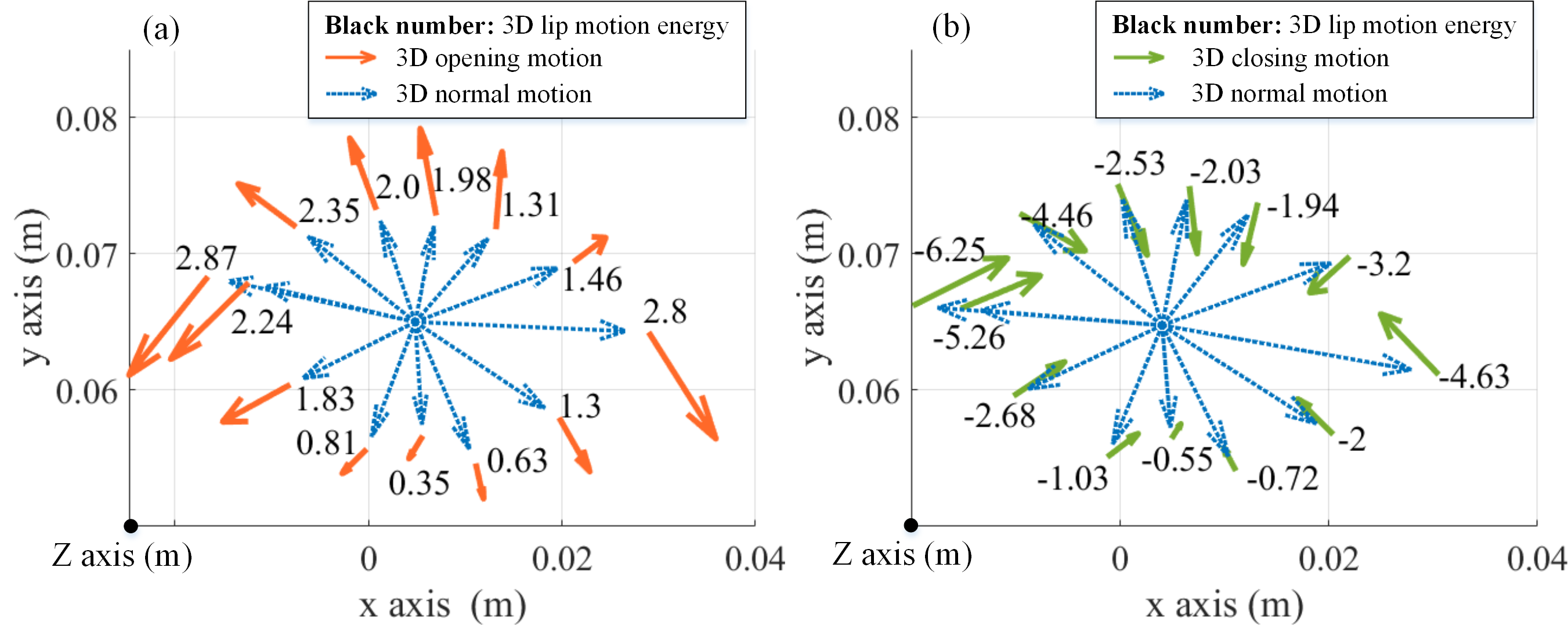}\\
  \caption{Motion divergence of 3D lip landmarks: (a) Lip opening event with red 3D motion vectors of lip landmarks, whose motion divergences are positive. The blue vectors are 3D reference vectors to the lip landmarks from the center of the reference sphere. (b) lip closing event with green 3D motion vectors of lip landmarks, whose motion divergences are negative.  (Note: for a better view, we show the 3D vectors in the XY view and scale the length of the vectors.)}
  \label{figdiver}
\end{figure}

\textbf{3D lip landmarks divergence}. We set the first frame $f_1$ and the last frame $f_{250}$ as initial reference frames for detecting the opening event and closing event, respectively. Since the video clip is synchronized with the audio clip, the first frame is earlier than the starting of the speech and the last frame covers the end of the speech. It is noted that the lip motion is sequential, so the motion state along the time axis is invariant to the reference time. The temporal detection resolutions are set as $\{30\times, 15\times, 7\times, 3\times 1\times\}$ original interframe interval (More analysis on temporal resolutions are presented in Sec. \ref{sec_multi-reso} below). At each temporal resolution, the proposed pipeline calculated the 3D interframe motion divergence along the time axis. Fig.\ref{figdiver} shows an example from a single frame of motion divergence of 3D lip landmarks at the temporal detection resolution of $30 \times$. We can see that the closing motion allows an angle of over 90 degrees with its reference motion vector (the blue vector), thus generating a negative motion divergence measure (Fig.\ref{figdiver}a), and vice versa (Fig.\ref{figdiver}b).

%

\subsection{Quantitative performance}
\subsubsection{Metrics}
We define three hierarchical metrics to evaluate both frame-level and event-level performance of the algorithm as below.

$\bullet$ Framewise Accuracy (F-Acc): our lip event detection issue is a sequence-to-sequence classification task. Each frame refers to one of the three activity states: static, opening, and closing. The F-Acc measures the classification rate of all the frame states.

$\bullet$ Event Frame Deviation (F-Dev): F-Dev evaluates the mean deviation between the detected events $T_i$ and corresponding ground truth events $gt_i$, as an event deviation $\delta = \left| T_i - gt_i \right|$. The deviation is caused by the event frames being detected earlier or later than the ground-truth frame.


$\bullet$ Event Recall Rate (E-RR): For a detected event, if its event frame deviation is within a F-Dev tolerance, the event is regarded as a true response. E-RR is the ratio of the correctly detected events to all of the events. The false events contain missing ones and incorrectly classified events, where the former is related to the magnitude of the motion energy, and the latter is caused by the wrong sign of 3D motion divergence. Thus, the E-RR increases with a larger F-Dev tolerance.

\begin{table*}
  \centering
  \caption{Multi-indictor comparison of different methods or configurations on the dataset}
  \begin{tabular}{lcccc}
  \hline
  Methods & F-Acc (\%) & F-Dev (opening/closing) & E-RR (\%) & T-Dev (ms)\\
  \hline
  Lip feature \cite{0051RF18} & 85.14 & 12.28/24.87 & 80.5 & 74.3\\
  TCN (2-fold CV) \cite{TCN} & 85.36 & 11.82/24.77 & 80.5 & 73.2 \\
  LSTM (2-fold CV) \cite{lstm} & 82.79 & 14.84/28.18 & 84.0 & 86.0 \\
  Single-scale (noisy features) & 82.20 & 14.01/30.5& 81.0 &89.0\\
  Single-scale (smoothed features)& 88.16 & \textbf{9.81}/19.78 & 88.0 & 59.2\\
  Ours (noisy features) & 87.58 & 14.08/16.98 & 90.0 & 62.1\\
  Ours (smoothed features)&\textbf{89.46} & 12.63/\textbf{13.72} & \textbf{91.5} & \textbf{52.7}\\
    \hline
  \multicolumn{5}{p{400pt}}{Note: F-Acc is Framewise Accuracy; F-Dev is Event Frame Deviation; E-RR is Event Recall Rate with the F-Dev tolerance of 40 frames; T-Dev is average Time Deviation of the events in frames converted to msec (frame per second is 250). All the indictors are mean values across the samples. CV is cross-validation.}
\end{tabular}
\label{table_compar}
\end{table*}

\subsubsection{Ablation study on temporal resolutions}
\begin{figure}
  \centering
  \includegraphics[width=8.2cm]{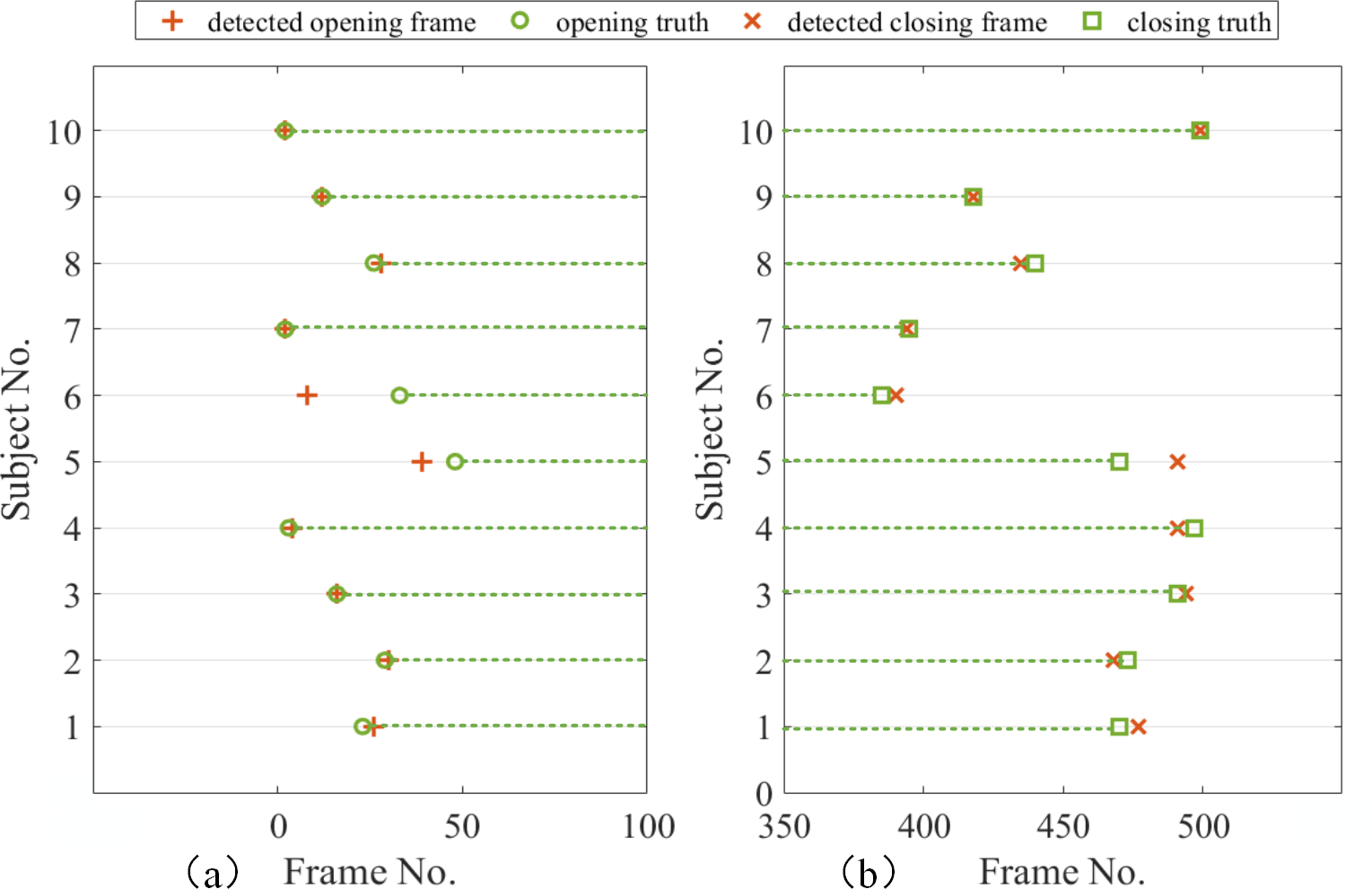}\\
  \caption{10 example results of lip event detection: (a) 3D lip opening; (b) 3D lip closing. The deviation between the detected moment and ground truth moment is measured by Event Frame Deviation (F-Dev). }\label{fig10dete}
\end{figure}

The temporal detection resolution is set to be 2 configurations: multi-temporal-resolutions and single frame resolution. We calculate multiple performance metrics for our algorithm across all the 100 test sequences (200 events including opening and closing). We test both configurations on noisy features and smoothed features (Sec.3.1) to investigate the robustness. Fig.\ref{fig10dete} presents 10 example results of true event detection by our algorithm on smoothed features. It shows that the opening and closing event frames detected are closely consistent with the ground truth. More comparative results are listed in the last four rows of Table \ref{table_compar} (the rest results are mentioned in the next section). We set the frame deviation tolerance as 40 frames (16 msec) and calculate an event recall rate (E-RR) for each method or configuration.

\begin{figure}
  \centering
  \includegraphics[width=8.1cm]{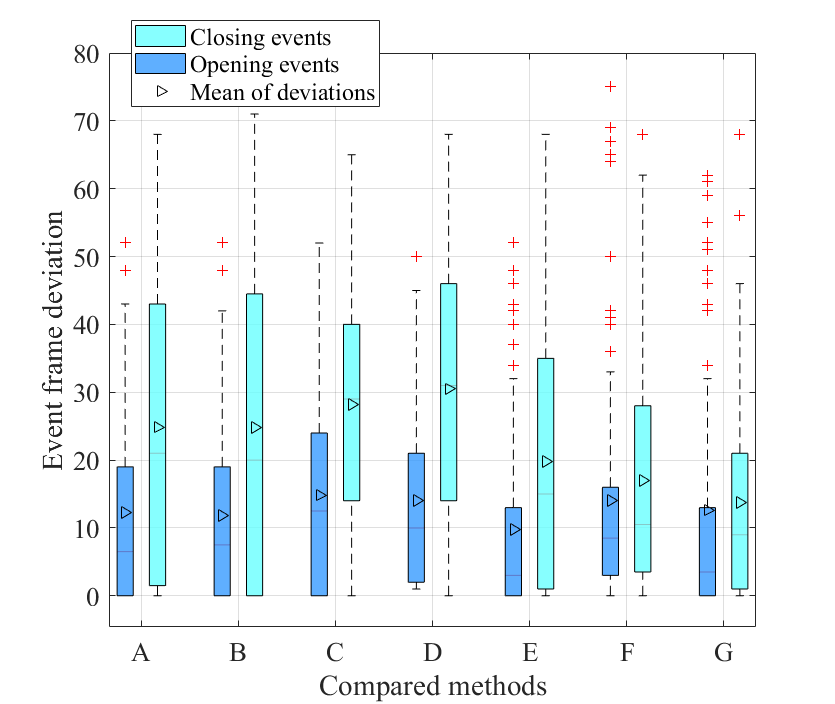}\\
  \caption{Event Frame deviation (F-Dev) of different methods or configurations: A is lip distance-based features \cite{ZhangF19}; B is LSTM \cite{lstm}; C is TCN \cite{TCN}; D is single-resolution on noisy features; E is single-resolution on smoothed features; F is multi-resolution on noisy features; G is multi-resolution on smoothed features. Each compared group consists of performance on opening events and closing events.}\label{fig_error}
\end{figure}

From the comparative results, we can see that our pipeline with multi-temporal-resolution on smoothed data achieves higher E-RR of $91.5\%$ and lower T-Dev of 52.7 msec. More detailed F-Dev results are shown in Fig.\ref{fig_error}. For the pipeline with multi-temporal-resolutions, the mean F-Dev. of opening and closing events are 12.6 frames and 13.7 frames, respectively. The multi-temporal resolution detection is more robust to noisy data, as the event proposal generated when using larger scale motion energy focuses the detection range.

\textbf{Failure cases analysis}. For the missed event samples, the motion energy was too small to be detected. The false detection samples are mainly caused by irregular lip motions and incorrect interframe motion. For the irregular lip motion, the lips of a few speakers went through some frames of deforming before opening and were somewhat widening while closing, which allows the lip landmarks to generate irregular motion divergence and thus influence the event decision. Besides, 3D lip landmark deviations or registration error will also influence the divergence feature, which could be improved by using more advanced lip landmark tracking or detection algorithms.


\subsection{Comparison on different methods}
We compare the performance of the proposed 3D lip detection algorithm with existing dynamic lip features and temporal detection methods on the same 3D point cloud streams. The comparison pipelines are 3D dynamic lip features \cite{ZhangF19} plus change time decision, divergence features plus temporal modeling method LSTM \cite{lstm}, and divergence features plus temporal modeling TCN \cite{TCN}. Both temporal models are trained under a 2-fold cross validation mode. The overall multi-indicator results are presented in Table 1, and Fig.8 shows the frame deviation distribution of all the samples. We also plot the E-RR vs. F-Dev tolerance curves in Fig.9 to indicate the correlation of the two metrics and the comprehensive performance of the compared methods.

We can see from the multi-aspect indicators that our pipeline with multi-temporal resolution outperforms others in terms of mean indicators. However, there are some outliers with large frame deviations (shown in Fig.8), which thus degrades the event recall rate when the F-Dev tolerance is larger than 60 frames (in Fig.9).

\begin{figure}
  \centering
  \includegraphics[width=8cm]{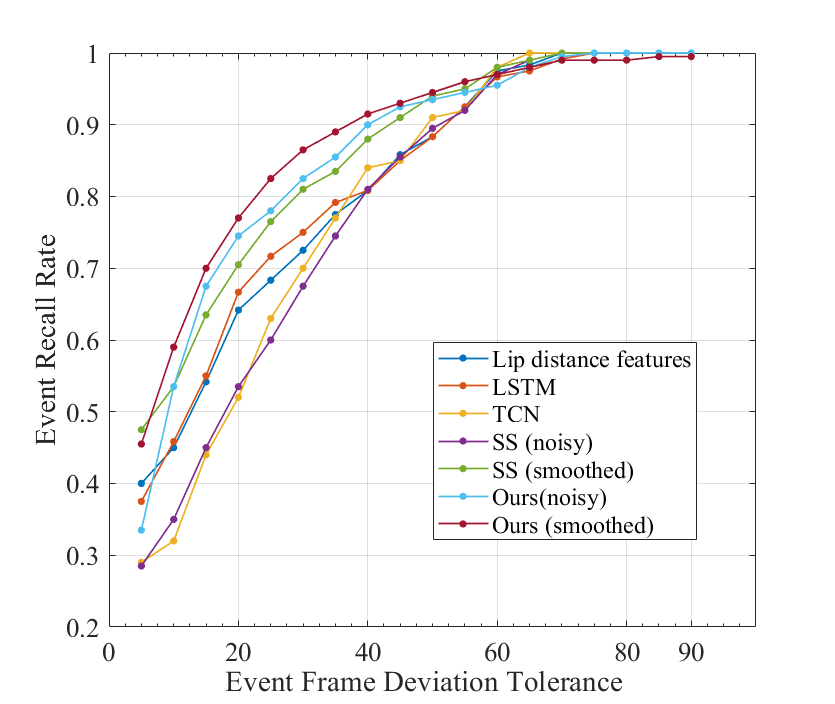}\\
  \caption{Event Recall Rate (E-RR) vs. varying framewise event deviation tolerance (F-Dev tolerance)}
  \label{fig_com}
\end{figure}


\subsection{Discussion on multi-temporal-resolutions}
\label{sec_multi-reso}
In our framework, $ts_t$ is an updated temporal resolution at which an event frame proposal is generated. We set the dynamic temporal resolution updated according to Eqn.\ref{eqn_effres}.
\begin{equation}
\label{eqn_effres}
ts_{t+1} = \lceil {\frac{ts_t}{k}} \rceil \;(ts_{t+1} \geq 1)
\end{equation}
where $k$ is an integral number that was set as $k=2$ in the experiments. The number $n$ of the temporal resolutions is related to the parameter $k$, satisfying $mod({\frac{ts_0}{k^(n-1)}})=0$.

At each temporal resolution $ts_t$, $gt_t$ is a relative ground truth frame updated with the reference frame according to Eqn.\ref{eqn_effupdate}.
\begin{equation}
\label{eqn_effupdate}
gt_{t+1} = w(gt_t-\lfloor {\frac{gt_t}{ts_t}} \rfloor ts_t) + (1-w)ts_t
\end{equation}
where $w\in[0,1]$ is a binary parameter that satisfies
\begin{equation}
\label{eqn_effw}
w = sgn[mod(gt_t,ts_t)]
\end{equation}

\textbf{Detection mode}. The event detection at each temporal resolution can be in a sequential or a parallel mode. In the sequential mode, the numbers of interframe event detections is related to the relative ground truth $gt_t$ and the resolution $ts_t$, as Eqn.\ref{eqn_eff}. The first item of Eqn.\ref{eqn_eff} indicates the number of interframe detections when the ground truth frame $gt_t$ is not integral multiples of the current temporal resolution $ts_t$, and the second item counts the number when $gt_t$ is integral multiples of $ts_t$.
\begin{equation}
\label{eqn_eff}
DetNum = \sum_{t=0}^nw(\lfloor {\frac{gt_t}{ts_t}} \rfloor + 1)+(1-w){\frac{gt_t}{ts_t}}
\end{equation}

\begin{figure}
\centering
  \includegraphics[width=8.2cm]{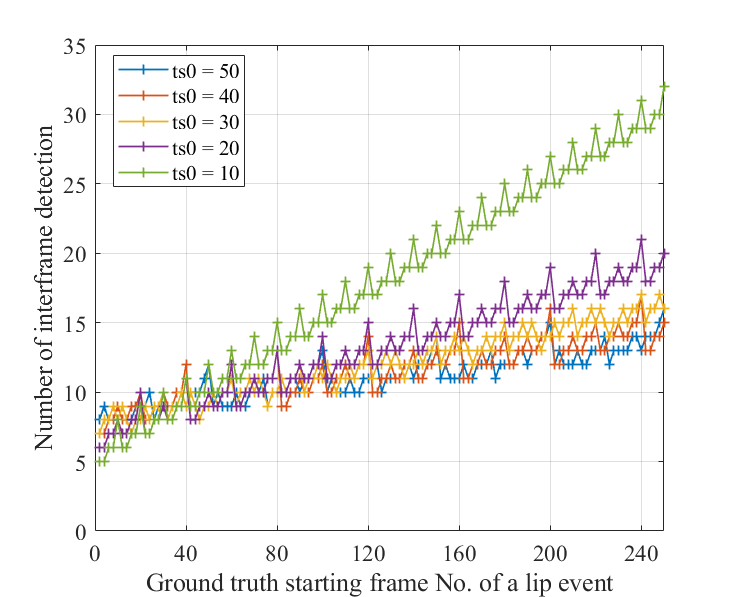}\\
  \caption{Number of interframe event detections for varying starting times of a lip event at different initial temporal resolutions $ts_0$ ($ts_0$ is measured by temporal interval of original resolution).}
  \label{figcurve}
\end{figure}

Based on the sequential mode analysis, we generated a set of synthetic ground truth event times $gt_0$ and initial temporal resolutions $ts_0$ to investigate how the number of interframe detections changes with both parameters. The results are shown in Fig.\ref{figcurve}. The lip event are a progressive motion, so the number of interframe detections changes with the event happening moment. A large initial temporal detection resolution saves the computational cost in the sequential interframe detection mode.

\subsection{Discussion on applicability}

\textbf{Various speaking speeds}. The test sequences exhibit different speaking speeds due to individual behaviors. Speaking slowly typically generates lower interframe motion energy. The interframe motion signature of lower motion energy is more likely to be influenced by data noise and inexact landmark localization. The multi-resolution strategy enlarges the interframe motion energy via an initially low detection resolution and reduces the search space gradually. For a slow speaker, a frame rate higher than that of the event is unnecessary. Thus, if an event is detected at a coarse resolution but not responded at the next finer resolution, our algorithm outputs the result at the coarse resolution as the event.

%

\textbf{Algorithm efficiency}. The time complexity of the pipeline is $O(n_{ts}f)$, which is related to the number of temporal resolutions $n_{ts}$ and the number of frames $f$ at one temporal resolution. If the interframe detection of a sequence at one temporal resolution is performed in a parallel mode, the computational complexity becomes $O(n_{ts})$. Besides, since the 3D data quality would influence the landmark extraction and pose correction, if higher-quality data acquisition allows the pipeline to be free from data denoising, the efficiency would be improved greatly. The pipeline is promising to be applied in real time.

\section{Conclusion}
This paper proposes a 3D lip event detection method at multiple temporal resolutions, with an interframe motion representation called 3D motion divergence. The method avoids the drawbacks of 2D intensity data and is a promising alternative for lip event detection. The experimental results demonstrate that the proposed pipeline achieves a state-of-the-art performance. The strategy of the multi-temporal resolution improves the robustness to various speaking speeds. The motion representation based on 3D lip landmarks avoids the complexity of the inner mouth. The 3D lip event pipeline helps automatically segment the 3D dynamic clips of interest and is beneficial for later global lip dynamics analysis.

In the future, we would like to focus on temporal modeling of 3D dynamic lips that represents both the interframe motion and the global dynamics, and seek to achieve 3D lip event detection and behavior analysis in an end-to-end fashion.

\section*{Acknowledgment}
The work is supported by Natural Science Foundation of China (NSFC) under Grant No.61906004.

\balance
{\small
\bibliographystyle{ieee_fullname}
\bibliography{ref}
}

\end{document}